\documentclass{article}

\PassOptionsToPackage{numbers, compress}{natbib}
\usepackage[dblblindworkshop, final]{neurips_2026}

\usepackage[utf8]{inputenc}
\usepackage[T1]{fontenc}
\usepackage{hyperref}
\usepackage{url}
\usepackage{booktabs}
\usepackage{amsfonts}
\usepackage{amsmath}
\usepackage{nicefrac}
\usepackage{microtype}
\usepackage{xcolor}
\usepackage{graphicx}
\usepackage{tikz}
\usetikzlibrary{positioning, arrows.meta, calc, matrix}

\title{Uncertainty, Not Concealment: A Mechanistic Reappraisal of Sycophancy in Vision-Language Models}

\workshoptitle{NeurIPS 2026 Workshop on Medical Reasoning with Vision Language Foundation Models}

\author{%
  \parbox[t]{0.28\textwidth}{\centering\small\textbf{Mohamed I.~Eldagla}\thanks{These authors contributed equally to this work.}\\[0.3em]\normalfont German International University in Cairo\\[0.1em]\normalfont mohamed.eldagla@\allowbreak giu-uni.de}%
  \And
  \parbox[t]{0.28\textwidth}{\centering\small\textbf{Zeyad H.~ElShafei}\footnotemark[1]\\[0.3em]\normalfont Arab Academy for Science, Technology and Maritime Transport\\[0.1em]\normalfont z.elshafei7151@\allowbreak student.aast.edu}%
  \And
  \parbox[t]{0.28\textwidth}{\centering\small\textbf{Ahmed M.~Anter}\\[0.3em]\normalfont Egypt-Japan University of Science and Technology\\[0.1em]\normalfont ahmed.anter@\allowbreak ejust.edu.eg}%
}

\begin{document}

\maketitle

\begin{abstract}
Vision-language models (VLMs) in medical retrieval-augmented generation must be able to say when a retrieved report contradicts the X-ray, so understanding whether they can detect such conflicts is important for safe deployment. The \emph{sycophancy mask} hypothesis holds that these models internally detect conflicts but refuse to report them, concealing a detected contradiction. We introduce a three-part diagnostic that separates \emph{concealment} from \emph{uncertainty}: representation probing, activation steering, and two-way forced choice. Across four instruction-tuned VLMs on chest X-rays, we confirm behavioral suppression of contradiction and find that a linear probe separates matched from mismatched inputs, beyond what a simple CLIP model achieves. Steering the learned conflict direction never elicits contradiction, and forcing a binary choice does not recover one; instead, cross-modal attention to the image drops on mismatched pairs. We conclude that these models fail to contradict because the mismatch registers as hedged, gated uncertainty rather than concealed knowledge, calling for re-grounding and uncertainty surfacing. Code is available at \url{https://github.com/mohamed-eldagla/Uncertainty-Not-Concealment/}.
\end{abstract}

\section{Introduction}

In clinical retrieval-augmented generation (RAG)~\citep{lewis2020rag}, a VLM is shown an X-ray and a retrieved report and asked to answer against the image. If the report is wrong, the model must \emph{contradict} it. A central safety question is \emph{epistemic}: when image and retrieved text conflict, does the model \emph{know} it is wrong and suppress that knowledge (concealment), or fail to resolve the conflict at all (uncertainty)?

Prior work on text-only language models establishes \emph{sycophancy}, a tendency to agree with a user's stated, possibly false, beliefs~\citep{perez2023discovering,sharma2024sycophancy}. A natural extension to VLMs is the \emph{sycophancy mask} hypothesis: the model internally \emph{detects} an image--text conflict but systematically \emph{refuses to report} it, emitting a $0\%$ contradiction rate on mismatched pairs. This reading follows directly from a single observation: an internal signal present while the behavioral signal is absent. But that observation is equally consistent with \emph{concealment} (a correct ``CONTRADICTS'' answer is masked) and with \emph{uncertainty} (the conflict is never resolved). Distinguishing the two is the subject of this paper, because it determines the intervention.

More broadly, whether internal representations diverge from stated answers is central to the faithfulness literature: unsupervised probes~\citep{burns2022ccs,kadavath2022know}, truth directions~\citep{marks2023geometry}, and representation editing~\citep{zou2023repeng} operate in text-only models, while VLM representations have been probed and edited for hallucination~\citep{jiang2025interpreting,liu2025mi}. Closest to our setting, refined activation steering~\citep{lfsteering2025,contextfocus2026} may find more effective directions than our coarse one; grounded medical pre-training targets cross-modal mismatch~\citep{gkmvlp2024}; and uncertainty-triggered dynamic retrieval~\citep{qucorag2026,dynamicrag2026} operationalizes uncertainty surfacing. Our contribution is the diagnostic: three tests separating concealment from uncertainty.

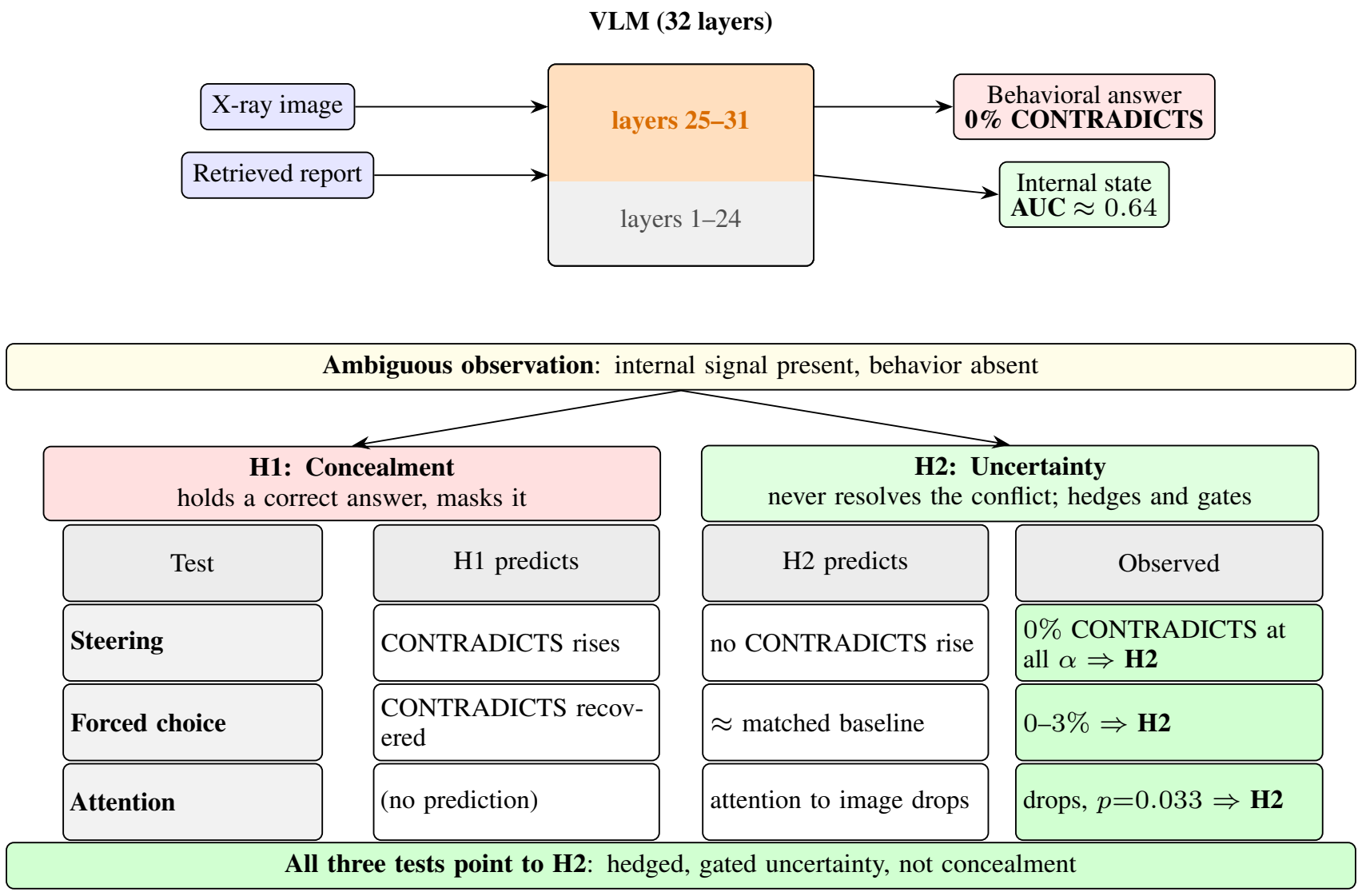
\begin{figure}[ht]
  \centering
  \resizebox{0.80\linewidth}{!}{%
  \begin{tikzpicture}[
    font=\scriptsize, >=Stealth,
    box/.style={draw, rounded corners=2pt, inner sep=3pt, align=center},
    inp/.style={box, fill=blue!10},
    mdl/.style={draw, rounded corners=2pt, fill=gray!12, outer sep=0pt},
    beh/.style={box, fill=red!10},
    int/.style={box, fill=green!10},
    hyp/.style={box, text width=5.6cm, inner sep=3pt},
    tst/.style={box, fill=gray!10, align=left, text width=2.3cm, inner sep=2pt, minimum height=0.72cm},
    pred/.style={box, align=left, text width=2.55cm, inner sep=2pt, minimum height=0.72cm},
    obs/.style={box, align=left, text width=2.75cm, inner sep=2pt, minimum height=0.72cm, fill=green!18},
    hdr/.style={box, fill=gray!15, text width=2.3cm, align=center, inner sep=2pt, minimum height=0.72cm},
  ]
  \node[mdl, minimum height=1.9cm, minimum width=2.5cm] (vlm) at (0,0) {};
  \node[font=\scriptsize\bfseries, above=1.5mm of vlm.north] {VLM (32 layers)};
  \begin{scope}
    \clip[rounded corners=2pt] (vlm.south west) rectangle (vlm.north east);
    \fill[orange!25] (vlm.north west) rectangle ($(vlm.north east)+(0,-1.1)$);
  \end{scope}
  \draw[draw=black, rounded corners=2pt] (vlm.south west) rectangle (vlm.north east);
  \node[font=\scriptsize\bfseries, orange!85!black] at ($(vlm.north)+(0,-0.55)$) {layers 25--31};
  \node[font=\scriptsize, gray!60!black] at ($(vlm.south)+(0,0.42)$) {layers 1--24};
  \node[inp] (img) at (-3.8,0.55) {X-ray image};
  \node[inp, below=2mm of img] (rep) {Retrieved report};
  \node[beh] (beh) at (3.8,0.55) {Behavioral answer\\[-0.2em]\textbf{0\% CONTRADICTS}};
  \node[int, below=2mm of beh] (int) {Internal state\\[-0.2em]\textbf{AUC $\approx 0.64$}};
  \draw[->] (img.east) -- (img.east -| vlm.west);
  \draw[->] (rep.east) -- (rep.east -| vlm.west);
  \draw[->] (vlm.east |- img.east) -- (beh.west);
  \draw[->] (vlm.east |- rep.east) -- (int.west);
  \node[box, fill=yellow!10, text width=12.5cm] (amb) at (0,-1.9) {\textbf{Ambiguous observation}: internal signal present, behavior absent};
  \node[hyp, fill=red!12] (h1) at (-3.1,-3.0) {\textbf{H1: Concealment}\\holds a correct answer, masks it};
  \node[hyp, fill=green!12] (h2) at (3.1,-3.0) {\textbf{H2: Uncertainty}\\never resolves the conflict; hedges and gates};
  \draw[->] (amb.south) -- (h1.north);
  \draw[->] (amb.south) -- (h2.north);
  \node[hdr] (ht) at (-4.6,-3.75) {Test};
  \node[hdr, text width=2.55cm] (h1c) at (-1.55,-3.75) {H1 predicts};
  \node[hdr, text width=2.55cm] (h2c) at (1.55,-3.75) {H2 predicts};
  \node[hdr, text width=2.75cm] (ho) at (4.6,-3.75) {Observed};
  \node[tst] (s) at (-4.6,-4.5) {\textbf{Steering}};
  \node[pred] (s1) at (-1.55,-4.5) {CONTRADICTS rises};
  \node[pred] (s2) at (1.55,-4.5) {no CONTRADICTS rise};
  \node[obs] (s3) at (4.6,-4.5) {$0\%$ CONTRADICTS at all $\alpha$ $\Rightarrow$ \textbf{H2}};
  \node[tst] (f) at (-4.6,-5.25) {\textbf{Forced choice}};
  \node[pred] (f1) at (-1.55,-5.25) {CONTRADICTS recovered};
  \node[pred] (f2) at (1.55,-5.25) {$\approx$ matched baseline};
  \node[obs] (f3) at (4.6,-5.25) {$0$--$3\%$ $\Rightarrow$ \textbf{H2}};
  \node[tst] (a) at (-4.6,-6.0) {\textbf{Attention}};
  \node[pred] (a1) at (-1.55,-6.0) {(no prediction)};
  \node[pred] (a2) at (1.55,-6.0) {attention to image drops};
  \node[obs] (a3) at (4.6,-6.0) {drops, $p{=}0.033$ $\Rightarrow$ \textbf{H2}};
  \node[box, fill=green!18, text width=12.5cm] (conc) at (0,-6.6) {\textbf{All three tests point to H2}: hedged, gated uncertainty, not concealment};
  \end{tikzpicture}%
  }
  \caption{The diagnostic framework. A single observation (internal signal present, behavioral CONTRADICTS absent) is ambiguous between concealment (H1) and uncertainty (H2). Three tests (steering, forced choice, and cross-modal attention) disambiguate them; in every case the observation matches H2.}
  \label{fig:framework}
\end{figure}

We test this distinction with three complementary probes (Figure~\ref{fig:framework}): (1) \textbf{representation}: does the internal state separate matched from mismatched inputs, and is it non-trivial?; (2) \textbf{steering}: does amplifying the learned conflict direction surface CONTRADICTS (concealment) or only hedging (uncertainty)?; (3) \textbf{forced choice}: does removing the hedge option recover a hidden contradiction? We apply this to four instruction-tuned VLMs (LLaVA-1.5-7B/13B~\citep{liu2023llava}, Qwen2-VL-7B~\citep{wang2024qwen2vl}, InternVL2-8B~\citep{chen2024internvl}) on a chest X-ray dataset~\citep{demner2016} with controlled matched/mismatched conditions. We find: suppression is real; the internal signal exists and is non-trivial; but it is \emph{uncertainty} (hedging + visual gating), not a concealed contradiction.

\section{Setup}

We use frontal chest X-rays from the Indiana University dataset~\citep{demner2016}. We construct 4 different conditions: \textbf{C0} matched (image + true report, $n{=}738$); \textbf{C2} normal image + abnormal report ($n{=}257$); \textbf{C3} abnormal image + normal report ($n{=}481$); \textbf{C4} abnormal image + most-similar normal report ($n{=}234$). We evaluate four instruction-tuned VLMs in 8-bit; mechanistic experiments use InternVL2-8B (full template fidelity). Models are prompted to answer with a two-line template: \texttt{VISUAL EVIDENCE:\ [SUPPORTS|CONTRADICTS|INCONCLUSIVE]} followed by a free-text \texttt{RATIONALE}. Verdicts are parsed with a \emph{robust} first-verdict rule that extracts the first \texttt{VISUAL EVIDENCE:} label (brackets optional) or a leading bare label and labels non-matching or template-echoing outputs \texttt{INVALID}; this avoids naive substring matching over the whole output, which may also include negated hedges in the rationale (``does not contradict'') as contradictions. Generation uses greedy decoding (temperature $0$, \texttt{do\_sample=False}, up to $80$ new tokens, seed $42$), so verdict differences are not a decoding artifact.

\emph{Probes.} \textbf{CLS probe}: logistic regression on the last-token hidden state at the penultimate layer, trained to separate C0 from C2/C3 (grouped cross-validation; C4 is evaluated behaviorally). \textbf{Steering (CAAS)}: a direction $d=\mu_{\text{mismatched}}-\mu_{\text{matched}}$ computed from \emph{baseline}-mode hidden states (layers 25--31), added to the residual stream during generation at strength $\alpha$~\citep{turner2023steering}. \textbf{Causal patching}: replace visual-token activations between matched and mismatched contexts~\citep{meng2022rome}. \textbf{Forced choice}: the same task with the INCONCLUSIVE option removed.

\section{Behavioral suppression is real}

Table~\ref{tab:verdicts} reports the robust verdict distribution. No model contradicts: the rate is $0\%$ in every cell. Template compliance ranges from $60.7$--$78.6\%$ INVALID for LLaVA-7B to $0\%$ for Qwen2-VL. Hedging is model- and condition-specific: InternVL2-8B INCONCLUSIVE is $14.8\%$ in C2 but $2.7\%$/$0.4\%$ in C3/C4; Qwen2-VL hedges in under $1\%$ of cases.

\begin{table}[!b]
  \caption{Behavioral verdicts (\%) by condition, robust first-verdict parsing with multi-label outputs labeled INVALID. CONTRADICTS is $0\%$ everywhere.}
  \label{tab:verdicts}
  \centering
  \small
  \resizebox{\linewidth}{!}{%
  \begin{tabular}{llrrrr}
    \toprule
    Model & CONDITION & SUPPORT & CONTRADICT & INCONCLUSIVE & INVALID \\
    \midrule
    LLaVA-7B & C2 / C3 / C4 & 21.4 / 38.5 / 39.3 & 0.0 / 0.0 / 0.0 & 0.0 / 0.0 / 0.0 & 78.6 / 61.5 / 60.7 \\
    InternVL2-8B & C2 / C3 / C4 & 79.0 / 97.3 / 99.1 & 0.0 / 0.0 / 0.0 & 14.8 / 2.7 / 0.4 & 6.2 / 0.0 / 0.4 \\
    Qwen2-VL & C2 / C3 / C4 & 99.6 / 99.2 / 100.0 & 0.0 / 0.0 / 0.0 & 0.4 / 0.8 / 0.0 & 0.0 / 0.0 / 0.0 \\
    LLaVA-13B & C2 / C3 / C4 & 49.4 / 74.4 / 68.4 & 0.0 / 0.0 / 0.0 & 0.0 / 0.0 / 0.0 & 50.6 / 25.6 / 31.6 \\
    \bottomrule
  \end{tabular}%
  }
\end{table}

\begin{table}[!b]
  \caption{CLS probe AUC for C0 vs C2+C3 (mean $\pm$ std over 5 folds); C4 excluded.}
  \label{tab:cls}
  \centering
  \begin{tabular}{lc}
    \toprule
    Model & AUC \\
    \midrule
    LLaVA-7B & $0.650 \pm 0.019$ \\
    InternVL2-8B & $0.638 \pm 0.031$ \\
    Qwen2-VL & $0.654 \pm 0.043$ \\
    LLaVA-13B & $0.647 \pm 0.013$ \\
    \bottomrule
  \end{tabular}
\end{table}

\section{The internal signal exists and is non-trivial}

A linear CLS probe separates matched C0 from mismatched C2+C3 inputs in every model (Table~\ref{tab:cls}), with grouped cross-validation so the matched and mismatched version of each image stay in the same fold. The signal is also \emph{non-trivial}, although the margin over CLIP is small: a plain CLIP model (BiomedCLIP~\citep{zhang2024biomedclip}) reaches image--text cosine AUC $0.531$ in the LLaVA control and $0.595$ on the W2 InternVL2 probe ($p{=}0.001$), while the VLM CLS probe reaches $0.638$.

\section{The signal is uncertainty, not a concealed answer}

\begin{figure}[t]
  \centering
  \includegraphics[width=0.52\linewidth]{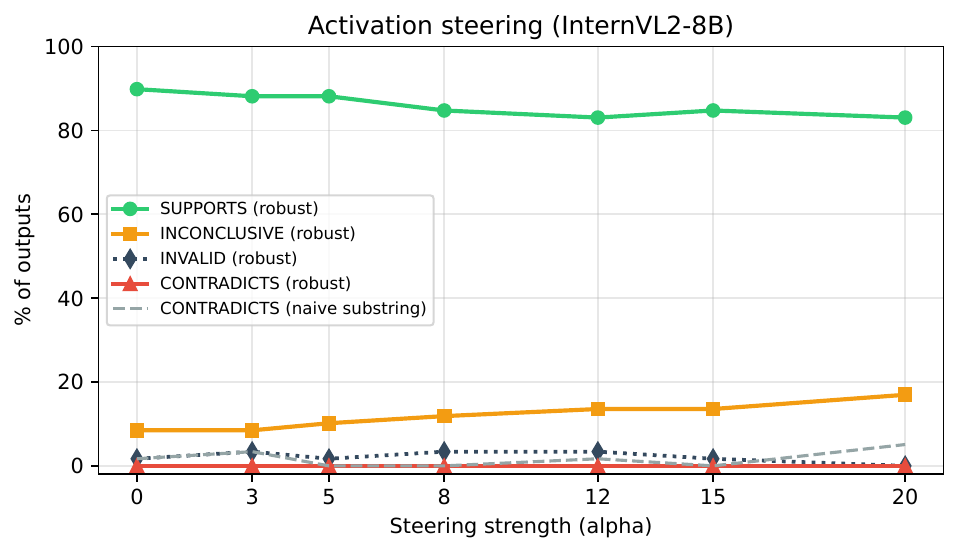}
  \caption{Activation steering (InternVL2-8B): robust CONTRADICTS stays at $0\%$ at every $\alpha$; SUPPORTS falls from $89.8\%$ to $83.1\%$ while INCONCLUSIVE rises from $8.5\%$ to $16.9\%$; INVALID remains $0$--$3.4\%$.}
  \label{fig:mech}
\end{figure}

\paragraph{Steering never elicits contradiction.} If the model concealed a ``CONTRADICTS'' answer, amplifying the conflict direction should surface it. It does not (Figure~\ref{fig:mech}): at every strength $\alpha\in\{0,3,5,8,12,15,20\}$ the robust CONTRADICTS rate stays at $0\%$. The SUPPORTS share declines from $89.8\%$ at $\alpha{=}0$ to $83.1\%$ at $\alpha{=}20$, with INCONCLUSIVE rising from $8.5\%$ to $16.9\%$; INVALID remains $0$--$3.4\%$. Naive ``contradict''-mentions (negated hedges such as ``does not contradict'') remain low ($0$--$5.1\%$) and never become robust contradictions. A false-positive control (steering matched pairs) yields $0\%$ robust CONTRADICTS.

\paragraph{The mismatch manifests as visual suppression.} Raw output entropy is uninformative (AUC $0.466$, $p{=}0.242$). The mechanistic signature is instead in \emph{attention}: cross-modal attention from text to visual tokens \emph{drops} on mismatched pairs (AUC $0.438$, $p{=}0.033$). The effect is modest but remains in the expected direction.

\paragraph{Causal patching shows a weak, mid-layer contribution.} Patching visual tokens between contexts shifts the mismatch readout in the expected direction with a clean no-op control ($\Delta{=}0.000$), but the effect is small, peaks at layer 20 ($\Delta P\approx{-}0.010$/$+0.015$), and vanishes by layer 30, consistent with gating, not an explicit contradiction.

\paragraph{Forced choice does not recover a hidden contradiction.} Removing the INCONCLUSIVE option (InternVL2-8B) does not recover a contradiction: C2, C3, and C4 yield CONTRADICTS at $3.0\%$, $0\%$, and $0\%$, versus $1.3\%$ on matched C0; C2 is only marginally above baseline. C4 yields $0/100$.

\section{Contributions}

\textbf{(1) A diagnostic framework separating concealment from uncertainty:} representation, steering, and forced choice jointly disambiguate an observation that is otherwise consistent with both.

\textbf{(2) Evidence for uncertainty-based sycophancy with a mechanistic signature:} steering never produces CONTRADICTS and forced choice does not recover one, while \emph{attention-gating} (reduced cross-modal attention on mismatch) defers to the text.

\textbf{(3) A caution against the ``mask'' framing:} the failure is uncertainty (hedging plus visual gating), not concealed knowledge, so the remedy is re-grounding and uncertainty surfacing, not forced disclosure.

\section{Limitations}

This work is limited to a single domain (chest X-rays) and one mechanistic model (InternVL2-8B); the attention evidence is correlational, the steering direction is coarse, and the causal-patching evidence is suggestive. The uncertainty interpretation is strongest for InternVL2-8B.

\section{Discussion and Future Work}

Our three tests converge on a single picture: the mismatch is reflected in the internal state; the models gate the visual signal and never express it as a contradiction, although the verbal form varies across models. The practical consequence is that interventions assuming a hidden correct answer, such as harder prompting or forced disclosure, will not recover a contradiction the model does not determinately hold; grounding interventions that keep visual attention high under conflict, and calibration or uncertainty-surfacing mechanisms that convert the internal signal into an explicit, graded statement, are more promising. Future work should verify the attention-gating signature across domains and models, add a capability ceiling check, isolate the cross-modal component of the CLS probe with a text-only ablation, and evaluate stronger steering methods~\citep{lfsteering2025,contextfocus2026} or a logit-based readout.

\bibliographystyle{unsrtnat}

\bibliography{references}

@inproceedings{lewis2020rag,
  author    = {Lewis, Patrick and Perez, Ethan and Piktus, Aleksandra and Petroni, Fabio and Karpukhin, Vladimir and Goyal, Naman and K{\"u}ttler, Heinrich and Lewis, Mike and Yih, Wen-tau and Rockt{\"a}schel, Tim and Riedel, Sebastian and Kiela, Douwe},
  title     = {Retrieval-Augmented Generation for Knowledge-Intensive {NLP} Tasks},
  booktitle = {Advances in Neural Information Processing Systems},
  volume    = {33},
  pages     = {9459--9474},
  publisher = {Curran Associates, Inc.},
  year      = {2020}
}

@article{demner2016,
  author  = {Demner-Fushman, Dina and Kohli, Marc D. and Rosenman, Marc B. and Shooshan, Sonya E. and Rodriguez, Laritza and Antani, Sameer and Thoma, George R. and McDonald, Clement J.},
  title   = {Preparing a Collection of Radiology Examinations for Distribution and Retrieval},
  journal = {Journal of the American Medical Informatics Association},
  volume  = {23},
  number  = {2},
  pages   = {304--310},
  year    = {2016},
  doi     = {10.1093/jamia/ocv080}
}

@inproceedings{liu2023llava,
  author    = {Liu, Haotian and Li, Chunyuan and Wu, Qingyang and Lee, Yong Jae},
  title     = {Visual Instruction Tuning},
  booktitle = {Advances in Neural Information Processing Systems},
  volume    = {36},
  pages     = {34892--34916},
  publisher = {Curran Associates, Inc.},
  year      = {2023},
  doi       = {10.52202/075280-1516}
}

@inproceedings{chen2024internvl,
  author    = {Chen, Zhe and Wu, Jiannan and Wang, Wenhai and Su, Weijie and Chen, Guo and Xing, Sen and Zhong, Muyan and Zhang, Qinglong and Zhu, Xizhou and Lu, Lewei and Li, Bin and Luo, Ping and Lu, Tong and Qiao, Yu and Dai, Jifeng},
  title     = {InternVL: Scaling up Vision Foundation Models and Aligning for Generic Visual-Linguistic Tasks},
  booktitle = {IEEE/CVF Conference on Computer Vision and Pattern Recognition (CVPR)},
  pages     = {24185--24198},
  year      = {2024}
}

@article{wang2024qwen2vl,
  author  = {Wang, Peng and Bai, Shuai and Tan, Sinan and Wang, Shijie and Fan, Zhihao and Bai, Junyang and Chen, Keqin and Liu, Xuejing and Wang, Jialin and Ge, Wenbin and Fan, Yang and Dang, Kai and Du, Mengfei and Ren, Xuancheng and Men, Rui and Liu, Dayiheng and Zhou, Chang and Zhou, Jingren and Lin, Junyang},
  title   = {Qwen2-VL: Enhancing Vision-Language Model's Perception of the World at Any Resolution},
  journal = {arXiv preprint arXiv:2409.12191},
  year    = {2024}
}

@article{zhang2024biomedclip,
  author  = {Zhang, Sheng and Xu, Yanbo and Usuyama, Naoto and Xu, Hanwen and Bagga, Jaspreet and Tinn, Robert and Preston, Sam and Rao, Rajesh and Wei, Mu and Valluri, Naveen and Wong, Cliff and Tupini, Andrea and Wang, Yu and Mazzola, Matt and Shukla, Swadheen and Liden, Lars and Gao, Jianfeng and Crabtree, Angela and Piening, Brian and Poon, Hoifung},
  title   = {A Multimodal Biomedical Foundation Model Trained from Fifteen Million Image-Text Pairs},
  journal = {NEJM AI},
  volume  = {2},
  year    = {2024},
  doi     = {10.1056/AIoa2400640}
}

@inproceedings{perez2023discovering,
  author    = {Perez, Ethan and Ringer, Sam and Luko\v{s}i\=ut\.e, Kamile and Nguyen, Karina and Chen, Edwin and Heiner, Scott and Pettit, Craig and Olsson, Catherine and Kundu, Sandipan and Kadavath, Saurav and Jones, Andy and Chen, Anna and Mann, Benjamin and Israel, Brian and Seethor, Bryan and McKinnon, Cameron and Olah, Christopher and Yan, Da and Amodei, Daniela and Amodei, Dario and Drain, Dawn and Li, Dustin and Tran-Johnson, Eli and Khundadze, Guro and Kernion, Jackson and Landis, James and Kerr, Jamie and Mueller, Jared and Hyun, Jeeyoon and Landau, Joshua and Ndousse, Kamal and Goldberg, Landon and Lovitt, Liane and Lucas, Martin and Sellitto, Michael and Zhang, Miranda and Kingsland, Neerav and Elhage, Nelson and Joseph, Nicholas and Mercado, Noemi and DasSarma, Nova and Rausch, Oliver and Larson, Robin and McCandlish, Sam and Johnston, Scott and Kravec, Shauna and El Showk, Sheer and Lanham, Tamera and Telleen-Lawton, Timothy and Brown, Tom and Henighan, Tom and Hume, Tristan and Bai, Yuntao and Hatfield-Dodds, Zac and Clark, Jack and Bowman, Samuel R. and Askell, Amanda and Grosse, Roger and Hernandez, Danny and Ganguli, Deep and Hubinger, Evan and Schiefer, Nicholas and Kaplan, Jared},
  title     = {Discovering Language Model Behaviors with Model-Written Evaluations},
  booktitle = {Findings of the Association for Computational Linguistics: ACL 2023},
  pages     = {13387--13434},
  publisher = {Association for Computational Linguistics},
  year      = {2023}
}

@inproceedings{sharma2024sycophancy,
  author    = {Sharma, Mrinank and Tong, Meg and Korbak, Tomek and Duvenaud, David and Askell, Amanda and Bowman, Sam and Durmus, Esin and Hatfield-Dodds, Zac and Johnston, Scott and Kravec, Shauna and Maxwell, Timothy and McCandlish, Sam and Ndousse, Kamal and Rausch, Oliver and Schiefer, Nicholas and Yan, Da and Zhang, Miranda and Perez, Ethan},
  title     = {Towards Understanding Sycophancy in Language Models},
  booktitle = {International Conference on Learning Representations},
  volume    = {2024},
  pages     = {110--144},
  year      = {2024}
}

@inproceedings{liu2025mi,
  author    = {Liu, Yiming and Zhang, Yuhui and Yeung-Levy, Serena},
  title     = {Mechanistic Interpretability Meets Vision Language Models: Insights and Limitations},
  booktitle = {The Fourth Blogpost Track at ICLR 2025},
  year      = {2025}
}

@inproceedings{jiang2025interpreting,
  author    = {Jiang, Nick and Kachinthaya, Anish and Petryk, Suzanne and Gandelsman, Yossi},
  title     = {Interpreting and Editing Vision-Language Representations to Mitigate Hallucinations},
  booktitle = {The Thirteenth International Conference on Learning Representations},
  year      = {2025}
}

@article{turner2023steering,
  author  = {Turner, Alexander Matt and Thiergart, Lisa and Leech, Gavin and Udell, David and Vazquez, Juan J. and Mini, Ulisse and MacDiarmid, Monte},
  title   = {Activation Addition: Steering Language Models Without Optimization},
  journal = {arXiv preprint arXiv:2308.10248},
  year    = {2023}
}

@inproceedings{meng2022rome,
  author    = {Meng, Kevin and Bau, David and Andonian, Alex and Belinkov, Yonatan},
  title     = {Locating and Editing Factual Associations in {GPT}},
  booktitle = {Advances in Neural Information Processing Systems},
  volume    = {35},
  pages     = {17359--17372},
  publisher = {Curran Associates, Inc.},
  year      = {2022},
  doi       = {10.52202/068431-1262}
}

@inproceedings{burns2022ccs,
  author    = {Burns, Collin and Ye, Haotian and Klein, Dan and Steinhardt, Jacob},
  title     = {Discovering Latent Knowledge in Language Models Without Supervision},
  booktitle = {The Eleventh International Conference on Learning Representations},
  year      = {2023}
}

@article{kadavath2022know,
  author  = {Kadavath, Saurav and Conerly, Tom and Askell, Amanda and Henighan, Tom and Drain, Dawn and Perez, Ethan and Schiefer, Nicholas and Hatfield-Dodds, Zac and DasSarma, Nova and Tran-Johnson, Eli and Johnston, Scott and El-Showk, Sheer and Jones, Andy and Elhage, Nelson and Hume, Tristan and Chen, Anna and Bai, Yuntao and Bowman, Sam and Fort, Stanislav and Ganguli, Deep and Hernandez, Danny and Jacobson, Josh and Kernion, Jackson and Kravec, Shauna and Lovitt, Liane and Ndousse, Kamal and Olsson, Catherine and Ringer, Sam and Amodei, Dario and Brown, Tom and Clark, Jack and Joseph, Nicholas and Mann, Ben and McCandlish, Sam and Olah, Chris and Kaplan, Jared},
  title   = {Language Models (Mostly) Know What They Know},
  journal = {arXiv preprint arXiv:2207.05221},
  year    = {2022}
}

@inproceedings{marks2023geometry,
  author    = {Marks, Samuel and Tegmark, Max},
  title     = {The Geometry of Truth: Emergent Linear Structure in Large Language Model Representations of True/False Datasets},
  booktitle = {First Conference on Language Modeling},
  year      = {2024}
}

@article{zou2023repeng,
  author  = {Zou, Andy and Phan, Long and Chen, Sarah and Campbell, James and Guo, Phillip and Ren, Richard and Pan, Alexander and Yin, Xuwang and Mazeika, Mantas and Dombrowski, Ann-Kathrin and Goel, Shashwat and Li, Nathaniel and Byun, Michael J. and Wang, Zifan and Mallen, Alex and Basart, Steven and Koyejo, Sanmi and Song, Dawn and Fredrikson, Matt and Kolter, J. Zico and Hendrycks, Dan},
  title   = {Representation Engineering: A Top-Down Approach to {AI} Transparency},
  journal = {arXiv preprint arXiv:2310.01405},
  year    = {2023}
}

@inproceedings{lfsteering2025,
  author    = {Yang, Jingyuan and Li, Rongjun and Wang, Weixuan and Zhou, Ziyu and Feng, Zhiyong and Peng, Wei},
  title     = {Latent Feature Activation Steering for Enhancing Semantic Consistency in Large Language Models},
  booktitle = {Natural Language Processing and Chinese Computing (NLPCC)},
  pages     = {168--180},
  publisher = {Springer},
  year      = {2025},
  doi       = {10.1007/978-981-95-3346-6_13}
}

@article{contextfocus2026,
  author  = {Anand, Nikhil and Somasundaram, Shwetha and Phukan, Anirudh and Saxena, Apoorv and Mukherjee, Koyel},
  title   = {ContextFocus: Activation Steering for Contextual Faithfulness in Large Language Models},
  journal = {arXiv preprint arXiv:2601.04131},
  year    = {2026}
}

@article{gkmvlp2024,
  author  = {Deng, Qiao and Huang, Zhongzhen and Wang, Yunqi and Wang, Zhichuan and Wang, Zhao and Zhang, Xiaofan and Dou, Qi and Hui, Yeung Yu and Hui, Edward S.},
  title   = {Grounded Knowledge-Enhanced Medical Vision-Language Pre-training for Chest X-Ray},
  journal = {Biomedical Signal Processing and Control},
  volume  = {112},
  pages   = {108416},
  year    = {2026},
  doi     = {10.1016/j.bspc.2025.108416}
}

@inproceedings{qucorag2026,
  author    = {Min, Dehai and Zhang, Kailin and Wu, Tongtong and Cheng, Lu},
  title     = {QuCo-RAG: Quantifying Uncertainty from the Pre-training Corpus for Dynamic Retrieval-Augmented Generation},
  booktitle = {Findings of the Association for Computational Linguistics: ACL 2026},
  pages     = {16482--16500},
  year      = {2026},
  doi       = {10.18653/v1/2026.findings-acl.812}
}

@inproceedings{dynamicrag2026,
  author    = {Li, Bo and Tian, Tian and Xu, Zhenghua and Cheng, Hao and Zhang, Shikun and Ye, Wei},
  title     = {Modeling Uncertainty Trends for Timely Retrieval in Dynamic {RAG}},
  booktitle = {Proceedings of the AAAI Conference on Artificial Intelligence},
  volume    = {40},
  pages     = {31527--31535},
  year      = {2026},
  doi       = {10.1609/aaai.v40i37.40418}
}

\end{document}